\documentclass[nohyperref]{article}

\usepackage{microtype}
\usepackage{graphicx}
\usepackage{subfigure}
\usepackage{booktabs} 

\usepackage{hyperref}

\usepackage[accepted]{icml2022}

\usepackage{amsmath}
\usepackage{amssymb}
\usepackage{mathtools}
\usepackage{amsthm}
\usepackage{xcolor}

\usepackage{graphicx}
\usepackage{verbatim}
\usepackage{natbib}
\usepackage{caption}

\usepackage[capitalize,noabbrev]{cleveref}

\theoremstyle{plain}

\theoremstyle{definition}

\theoremstyle{remark}

\usepackage[textsize=tiny]{todonotes}

\DeclareFontFamily{U}{matha}{\hyphenchar\font45}
\DeclareFontShape{U}{matha}{m}{n}{
      <5> <6> <7> <8> <9> <10> gen * matha
      <10.95> matha10 <12> <14.4> <17.28> <20.74> <24.88> matha12
      }{}
\DeclareSymbolFont{matha}{U}{matha}{m}{n}

\DeclareMathSymbol{\Lt}{3}{matha}{"CE}
\DeclareMathSymbol{\Gt}{3}{matha}{"CF}

\icmltitlerunning{Submission and Formatting Instructions for ICML 2022}

\begin{document}

\twocolumn[
\icmltitle{On the Subspace Structure of Gradient-Based Meta-Learning}



\icmlsetsymbol{equal}{*}

\begin{icmlauthorlist}
\icmlauthor{Gustaf Tegnér}{equal,yyy}
\icmlauthor{Alfredo Reichlin}{equal,yyy}
\icmlauthor{Hang Yin}{yyy}
\icmlauthor{Mårten Björkman}{yyy}
\icmlauthor{Danica Kragic}{yyy}

\end{icmlauthorlist}

\icmlaffiliation{yyy}{Division of Robotics, Perception and Learning, EECS, KTH Royal Institute of Technology, Stockholm, Sweden}

\icmlcorrespondingauthor{Gustaf Tegnér}{gustafte@kth.se}
\icmlcorrespondingauthor{Alfredo Reichlin}{alfrei@kth.se}

\icmlkeywords{Machine Learning, ICML}

\vskip 0.3in
]



\printAffiliationsAndNotice{\icmlEqualContribution} 

\begin{abstract}

In this work we provide an analysis of the distribution of the post-adaptation parameters of Gradient-Based Meta-Learning (GBML) methods. Previous work has noticed how, for the case of image-classification, this adaptation only takes place on the last layers of the network. We propose the more general notion that parameters are updated over a low-dimensional \emph{subspace} of the same dimensionality as the task-space and show that this holds for regression as well. Furthermore, the induced subspace structure provides a method to estimate the intrinsic dimension of the space of tasks of common few-shot learning datasets.

\end{abstract}

\section{Introduction}

Humans possess an innate ability to draw upon past experiences to efficiently solve new problems. From being exposed to multiple environments, we are able to learn general concepts which can be leveraged to solve tasks in novel domains. This concept of \emph{transfer} between problem domains is a core tenant of human intelligence. To endow this ability on intelligent learning systems, one can adhere to Multi-Task Learning (MTL), a paradigm that enables learning over multiple frameworks concurrently. A particular instance of MTL is Meta-Learning. These set of algorithms are designed to quickly learn and adapt to novel tasks. In general, this is done by exposing the learner to several tasks concurrently, allowing it to learn a common source of information between the tasks, as well as their differences.
\begin{figure}
    \centering
    \includegraphics[width=0.7\linewidth]{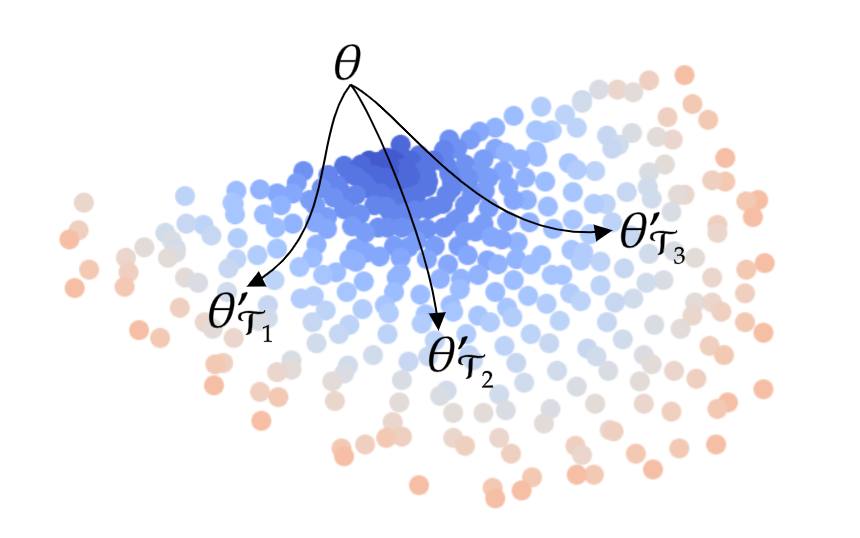}
    \caption{The space of task-adapted parameters for a sine regression task embedded in two-dimensional space. The polar-coordinate structure of the task is preserved in the space of task-adapted parameters.}
    \label{fig:wings}
\end{figure}
This ability to discriminate between tasks is at the core of meta learning models. A single task can be arbitrarily complex while a set of related tasks, instead, can form a well defined structure. In most cases, such a structure may be defined by only a small number of variables. That is, the space of related tasks can generally be assumed to be equipped with a very low \emph{intrinsic} dimension. Exploiting effectively this structure enables the meta learner to make an efficient use of its acquired knowledge. Adapting to novel tasks then essentially amounts to finding and exploiting these intrinsic factors of variation in the task space.

To give an example of such a structure, imagine the regression problem of meta-learning a collection of tasks $\{\mathcal{T}_i\}_{i=1}^N$ defined as the sum of sine waves of varying amplitude and phase. Learning a single task $\mathcal{T}_i$ can be made arbitrarily complex by considering additional sine waves. However, when learning multiple tasks in conjunction it suffices to identify the common form of the tasks and then learning only the relatively simple task structure induced by the varying task parameters. Figure \ref{fig:wings} depicts such a structure learnt by GBML. As shown, the task structure is \emph{embedded} in the task-adapted parameters themselves. 

In this paper we aim to study more explicitly this task structure through the lens of Gradient-Based Meta-Learning (GBML) \cite{finn2017model}. This particular family of meta learning methods adapt to a new task by learning how to generate a set of optimal parameters specific to that task. Recent work has shown how, in the context of classification, GBML requires to update only a subset of its parameters \cite{raghu2019rapid, oh2020boil}. We expand on this perspective by demonstrating that GBML implicitly learns a subspace that reflects the inherent structure of the tasks for both classification and regression. To investigate this phenomena, we borrow notions from dimensionality estimation and study the distribution of the parameters found by GBML after adaptation for each one of the tasks considered. By representing a task as the parameters of a task-adapted neural network, we find a high-dimensional representation of it. We can then leverage representation learning techniques such as dimensionality reduction to estimate the intrinsic dimensionality of the task space. We empirically show that GBML finds a subspace of a dimensionality that is optimal for the set of tasks at hand. This, in turn, provides us with an insight into the inherent subspace structure of common datasets.

Our contributions are as follows:
\begin{itemize}
    \item Empirical evidence that GBML methods naturally tend to a solution where the adapted parameters lie on a subspace of the smallest dimensionality possible given the meta-dataset.
    \item A method to analyze the dimensionality of the task-space of common datasets.
\end{itemize}

\section{Related Work}

\textbf{Meta-Learning}.
The general idea of meta-learning tackles the setting of learning a new task given only a small set of labeled examples. To achieve this, one of the most prominent approaches is model agnostic meta learning (MAML) \cite{finn2017model}. It learns a common parameter initialization across tasks that lies close to the optimal solution. Extensions to MAML couple the optimization procedure with a learned set of parameters to increase its expressivity \cite{lee2018gradient, flennerhag2019meta, park2019meta}.

\textbf{Few-shot Classification}. GBML for image classification has been extensively studied. \cite{raghu2019rapid} showed that GBML models naturally converge to the solution of adapting only the parameters of the last layer of an image classifier for $N$-way $K$-shot classification. They suggest that the test performances are correlated with the quality of the representation learner represented by the first part of the network. This has been confirmed by \cite{oh2020boil} where they also propose to freeze the adaptation of the last layer to incentivize the meta learner to find a solution that makes use of more layers. Further analysis have compared the behavior of MAML with a contrastive learning framework where the learned features exhibit invariance to the samples used in adaptation \cite{kao2021maml, goldblum2020unraveling, li2018learning}. 

\textbf{Dimensionality Reduction} A common accepted notion in machine learning is that high-dimensional data lies on a sub-manifold of a much lower dimension than the dimension of the ambient space. Dimensionality reduction techniques can be utilized to find this intrinsic dimensionality of data. Examples of such techniques are Principal Component Analysis (PCA) (Pearson, 1901) and Isomap (Tenenbaum et al., 2000). Isomap is a non-linear dimensionality reduction technique that aims to preserve geodesics of the original space in a space of dimension $k$. It works by minimizing the reconstruction error which is the difference between distances in the original space and embedded space w.r.t to the Isomap kernel. Previous work has utilized such techniques in an attempt to estimate the true intrinsic dimensionality of images \cite{pope2021intrinsic}. 

\textbf{Subspace Structure}.
Several works have investigated the geometry of the loss-landscape of neural networks \cite{li2018visualizing} \cite{lengyel2020genni}. In particular, it has been shown that there exists paths in the parameter space on which the loss remains low \cite{wortsman2021learning} \cite{garipov2018loss} \cite{frankle2020revisiting}. Other type of structure, such as hyper-planes has been explored in \cite{lengyel2020genni}. Recently, subspace structure has been considered in online reinforcement-learning by learning a subspace of policies which enables better generalization during testing \cite{gaya2021learning}. In this work, we extend this analysis to the meta-learning setting by identifying sub-manifold structure in the learnt parameter space.

\section{Preliminaries}

The general formulation of gradient-based meta-learning follows from \cite{finn2017model}. We consider a set of tasks $\mathcal{T}_i$ sampled from a distribution $p(\mathcal{T})$ together with a parameterized function $f_{\theta}$. 
Given a new task $\mathcal{T}_i$, GBML seeks to find a set of parameters $\theta$ such that applying one gradient step on the loss computed on a few data-points ($D^S$) of the task results in the optimal set of parameters $\theta'_{\mathcal{T}_i}$:
\begin{equation}
    \theta'_{\mathcal{T}_i} = \theta - \alpha \nabla_{\theta}\mathcal{L}_{\mathcal{T}_i}\left(f_\theta, D^S\right)
\end{equation}
To find such $\theta$, GBML optimizes the expectation of the loss, computed on the rest of the task's data ($D^Q$), of the model $f_{\theta}$ over all the tasks after adaptation:
\begin{equation}
    \mathcal{L}_{\text{GBML}} = \mathbb{E}_{p(\mathcal{T})} \left[\mathcal{L}_{\mathcal{T}}\left(f_{\theta - \alpha \nabla_{\theta}\mathcal{L}_{\mathcal{T}}(f_{\theta}, D^S) }, D^Q\right)\right].
\end{equation}

Let $S_{\mathcal{T}} = \text{supp}(\mathcal{T})$ be the support of the distribution of tasks. GBML essentially learns a differentiable map $M:S_{\mathcal{T}} \to \Theta$ from the support to the space of parameters of a neural network. Let $\Theta_{\mathcal{T}} = M(S_{\mathcal{T}})$ denote the image of this map, that is the space $\Theta_{\mathcal{T}} \subseteq \Theta$ induced by the task-adapted parameters found through GBML. Since $M$ is differentiable, we have that $\text{dim}(\Theta_{\mathcal{T}}) \leq \text{dim}(S_{\mathcal{T}})$. Assuming the tasks carry no redundancy, the dimensions are expected to be equal. The intrinsic complexity of the task can then be reduced to the study of the properties of $\Theta_{\mathcal{T}}$. 

\section{Evaluation Method}
We aim to investigate the geometry of the parametric model space of gradient-based meta-learning methods for different meta-problems. We hypothesize that since the space of tasks holds an intrinsic dimension $d$, our task-adapted parametric model space should be embedded in low-dimensional space of the same dimension. To investigate this hypothesis, we make use of dimensionality reduction techniques. Given a set of task-adapted parameters $\theta_{\mathcal{T}_i} \in \Theta= \mathbb{R}^D$, we make the assumption that $\theta_{\mathcal{T}_i}$ is sampled from a lower-dimensional manifold $\mathcal{M} \subseteq \mathbb{R}^D$ of intrinsic dimension $\text{dim}(\mathcal{M}) = d \ll D$. A common way to estimate the intrinsic dimensionality is to reduce the dimensions of $\theta_{\mathcal{T}_i}$ to a varying number of dimensions $k$. The intrinsic dimensionality would then correspond to the smallest $k$ which still preserves the features of the original data space, measured by the reconstruction error. We utilize Principal Component Analysis (PCA) \cite{pearson1901liii} and Isomap \cite{tenenbaum2000global} to estimate this intrinsic dimensionality.

\section{Experiments}

We conduct a number of experiments to analyze the parameter space that emerges when applying GBML to a number of regression and classification problems. For our experiments we used Model Agnostic Meta Learning (MAML) \cite{finn2017model}, one the most prominent members of the GBML family. To confirm that a structure of the parameters emerges independently of the expressiveness of the learner, we also used in the experiments Meta-Curvature (MC) \cite{park2019meta}. MC relies on learning a matrix $G$ which conditions the gradients by $G\nabla \mathcal{L}_{\theta}$. We perform our analysis on a toy classification and regression task and furthermore investigate few-shot classification on miniImagenet and Omniglot datasets. 
For the toy experiments we consider a two layer neural network with ReLU activations and $40$ hidden units. For miniImagenet and Omniglot, we use a $4$ layer convolutional neural network with batch-normalization, ReLU activations and max-pooling. We train the models for $100$ epochs with a meta-batch size of $32$. For Isomap, we use a neighborhood size of $20$.

\subsection{Classification Task}
In few-shot classification, it has been noted that GBML algorithms find a solution where only the last layers of the meta-learner gets updated during adaptation \cite{raghu2019rapid}. In our first experiment, we expand upon this insight by studying the entire space of task-adapted parameters for a toy $N$-way $K$-shot classification problem. To create this task, we construct a set of prototype classes $x_p \in \mathbb{R}^D$ that are equally spaced on a $D$-dimensional grid. We sample uniformly a subset of $N$ prototypes, each assigned to a distinct class $c \in [1, N]$. From each prototype, we now sample a set of $K$ examples $x_p^i \sim \mathcal{N}(x_p, \sigma)$. Adapting to a new task now involves changing the decision boundaries of a multinomial classification problem. We consider $5$ different datasets with the number of classes $N \in \{3,4,8,16\}$. To show that our method is architecture agnostic, we train our model by varying the number of hidden units in the last layer by $\{32, 64, 128 ,256 \}$. 

\begin{figure}
    \centering
    \includegraphics[width=0.45\textwidth]{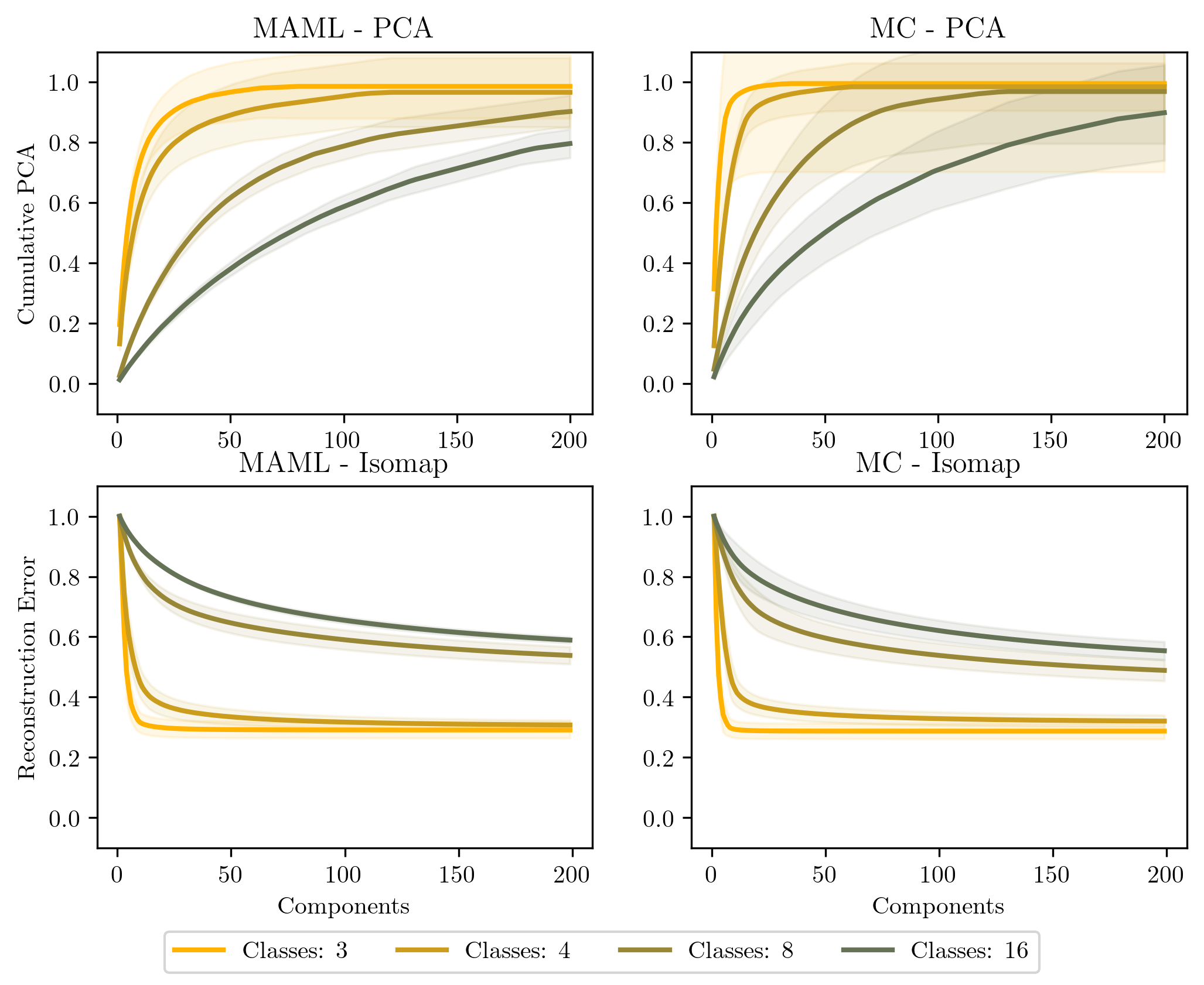}
    \caption{PCA and Isomap results for MAML and Meta-Curvature on the toy classification experiment.}
    \label{fig:toy_isomap}
\end{figure}
Since we are interested in identifying the intrinsic dimension $k$ from which no further increase in performance can be found, we normalize the Isomap scores by dividing them by the first (largest) value.
The plots are shown in Figure \ref{fig:toy_isomap}. As the number of classes increase, the number of dimensions required for adaptation increase accordingly. For PCA, it can be seen that for lower number of classes, more weight is put on a single principal component. This is reflected in the Isomap results where using a lower number of classes correspond to a steeper curve and thus lower dimension. 
Furthermore, the results are consistent with different architectures as the error bars remain intact. This shows that subspace structure is agnostic to the model used and is entirely dependent on the task itself. We can note that for two classes, a one-dimensional embedded task space seems to be the predominant result. In the appendix, Figure \ref{fig:mean_absolute_parameter_change} confirms that only a small subset of parameters in the last layers are updated for the binary classification problem while for other classes we note a larger change. While Figure \ref{fig:mean_absolute_parameter_change} re-confirms the results of \cite{raghu2019rapid}, we raise the point that it is not only what \emph{subset} of parameters are updated, but what \emph{subspace} they lie on. Although a subset of parameters alters with choice of architecture, the subspace is instead only dependent on the complexity of the problem itself.

\subsection{Regression Task}
Next we investigate if the same results holds true for regression tasks as well. 
We construct a regression task by considering the sum of sine waves with different amplitudes. To put it explicitly, for each task we sample a set of amplitudes $A_k \sim \mathcal{U}([0.1, 5.0])$ for $k=1,\ldots,N$ and construct a regression task as the sum:
\begin{equation}\label{eq:sine_experiment}
    y = \sum_{k = 1}^N A_k\sin(kx).
\end{equation}
Here, the number of sines $N$ defines the intrinsic dimension of the task space as the tasks vary only by the parameters $A_k$. For our experiments, we consider $N \in \{2,3,4,5\}$. We investigate the explained variance of PCA to compute an estimate of the dimensionality of the parametric model space. The results for PCA are shown in Figure \ref{fig:sine_results} while the ones for Isomap are included in the appendix (Figure \ref{fig:sine_isomap}).
\begin{figure}
    \centering
    \includegraphics[width=0.45\textwidth]{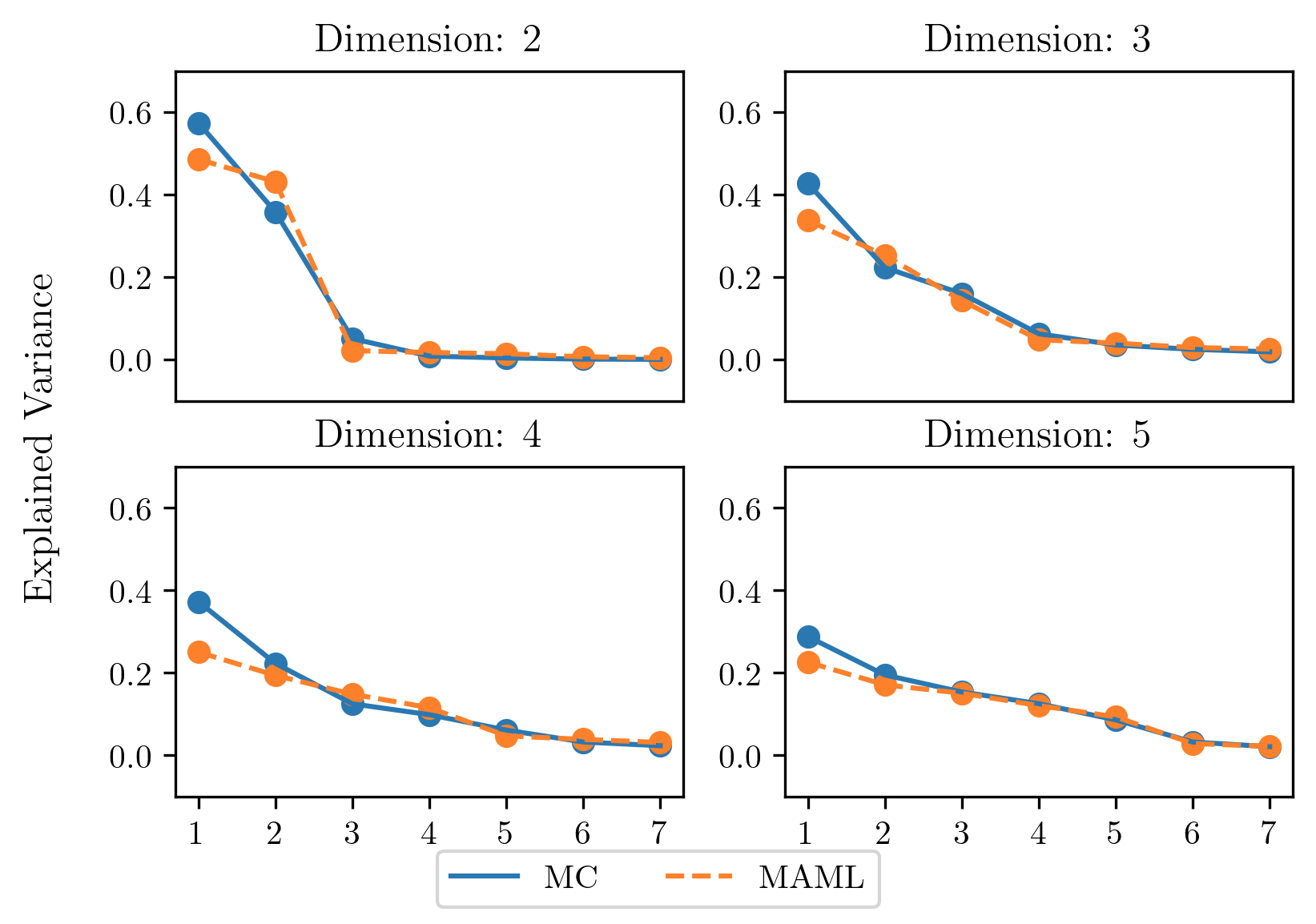}
    \caption{Explained variance of the first $k$ principal components of the task-adapted parameters of the sine tasks. PCA correctly identifies that the number of significant components is correlated with the intrinsic dimensionality of the task.}
    \label{fig:sine_results}
\end{figure}
From the figure, the strongest principal components coincide with the intrinsic dimensionality of the task. The effect declines as the complexity of the task increases, since there is an increased ambiguity for the meta-learner to learn complex functions with a limited amount of support data. This confirms our hypothesis that the intrinsic dimensionality of the task-space is preserved in the space of parameters induced by the meta-learner. 

\subsection{Analysis of Common Datasets}
\begin{figure}
    \centering
    \includegraphics[width=0.45\textwidth]{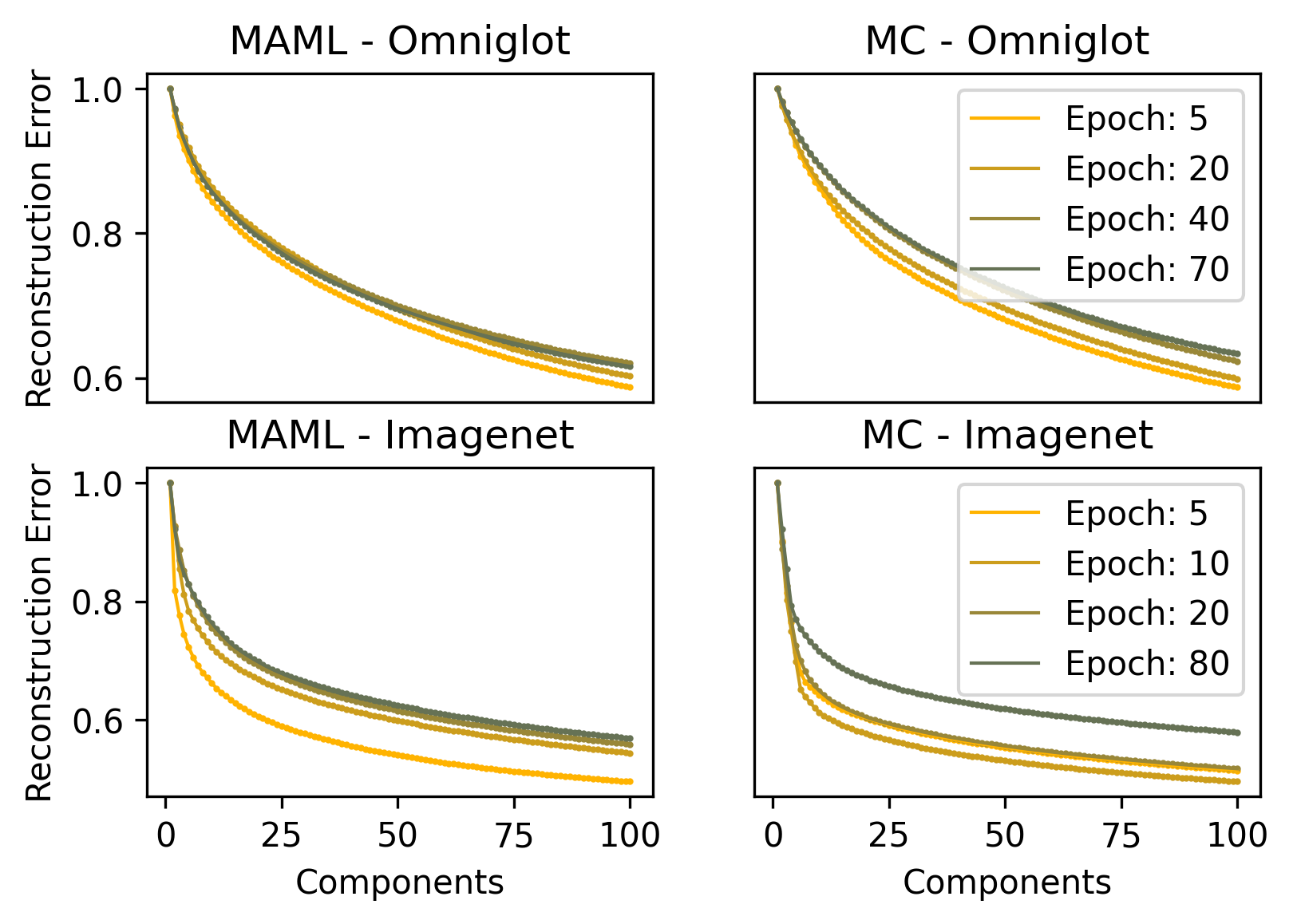}
    \caption{Estimated dimensionality from Isomap of Omniglot and miniImagenet. }
    \label{fig:Datasets_isomap}
\end{figure}

We now turn to a more complex setting of estimating the intrinsic dimensionality of common classification tasks. We consider an $N$-way, $K$-shot few shot learning setting with data from miniImagenet and Omniglot. For both of these tasks we let $N=5$ and $K=5$ and train our model for $100$ epochs. For the architecture, we consider the convolutional architecture specified in \cite{vinyals2016matching}.
Figure \ref{fig:Datasets_isomap} depicts the reconstruction error of Isomap for MAML and Meta-Curvature for both the miniImagenet and Omniglot tasks. The dimensionality reduction is performed on different epochs throughout the training. The results for miniImagenet show both a higher variability during training and a seemingly lower estimated dimensionality than Omniglot which is reflected in their performance on the test set as well. In the appendix we elaborate upon this observation.

\subsection{Reconstruction Experiment}
\begin{figure}
    \centering
    \includegraphics[width=0.3\textwidth]{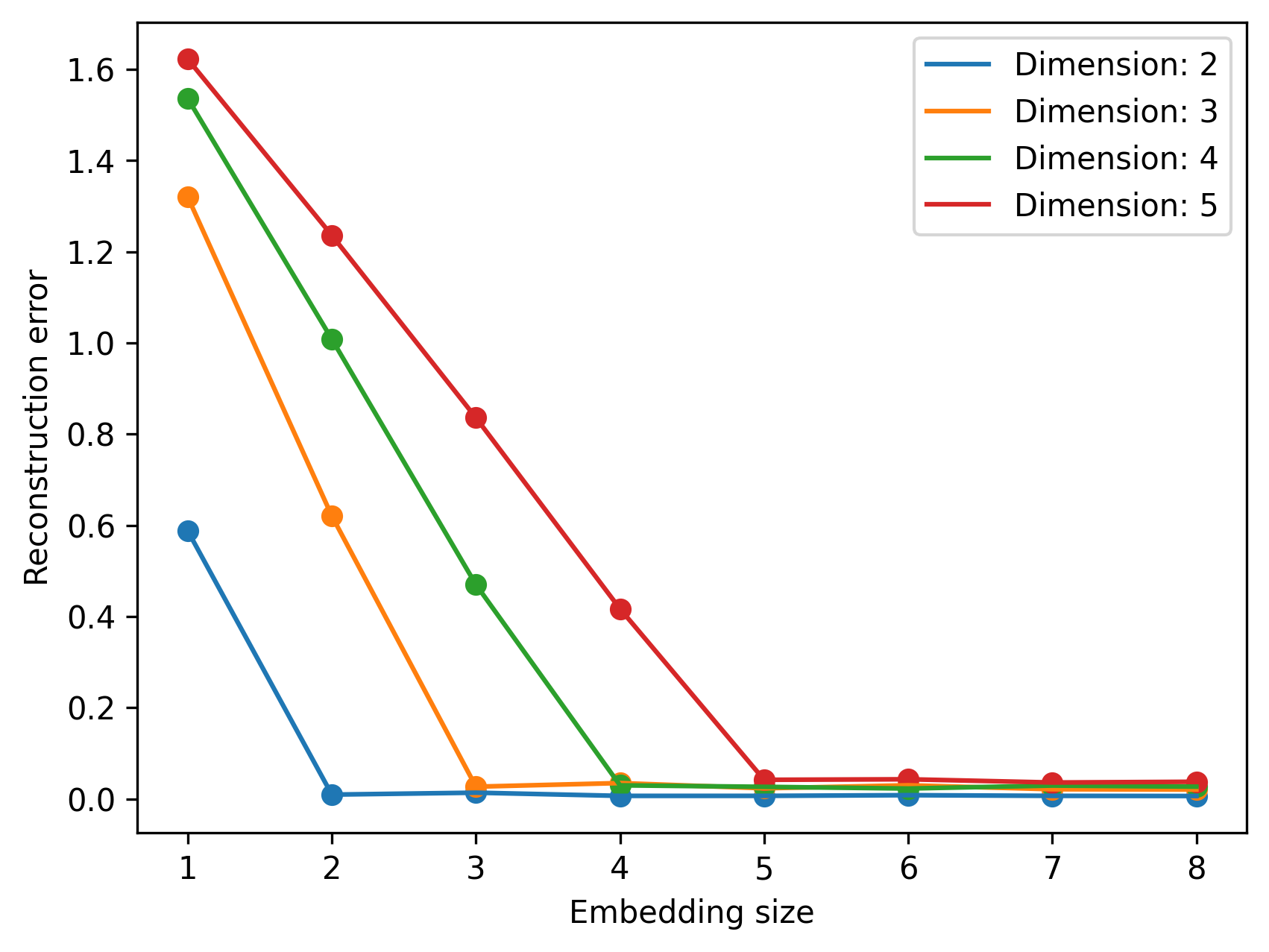}
    \caption{Regressing the task-parameters of the sinusoid task from the PCA embedding of the adapted parameters. The embedding size required to reach a low error is exactly the intrinsic dimension of the task.}
    \label{fig:sine_reconstruction}
\end{figure}

To further substantiate our findings, we perform an experiment involving reconstructing the task from the task-adapted parameters. We perform dimensionality reduction using PCA on the task-adapted parameters to find a representation $z \in \mathcal{Z}$ of reduced dimension $k$. From this we train a learner $f : \mathcal{Z} \to \mathcal{T}$ in a supervised manner to reconstruct the task from the low-dimensional embedding. For the sine-experiment, we attempt to regress the amplitudes defined in Equation $\ref{eq:sine_experiment}$. 
For the classification experiments, we do not have access to the ground-truth task-parameters. In this case, we consider classifying an image from a specific task given the embedding of the learnt parameters for that task. We perform the experiments for varying dimension $k$ and evaluate the results by considering the regression or classification performance. The results for the sine experiment are shown in Figure \ref{fig:sine_reconstruction}. As can be seen, the performance does not increase further as you increase the embedding size beyond the intrinsic dimension of the task. We evaluate the performance of $3, 8$ and $16$ classes for the toy-classification experiment. The results can be seen in Figure \ref{fig:toy_classification_reconstruction} in the Appendix. However, we argue that the discrete nature of classification tasks poses a limit in the analysis of the dimensionality of the manifold.

\section{Conclusion and Future Work}

In this work we proposed a method for analyzing the intrinsic dimensionality of the task-space learnt by gradient-based meta-learning methods. We provided an empirical analysis that reveals that GBML learn a low-dimensional subspace on which it performs adaptation on. Furthermore, we related this observation to the generalization performance of the model. We believe our analysis adds a valuable contribution to recent work that employ subspace structure in GBML. A possible future line of work is applying our method on a larger set of models and datasets in order to gain further insight. As a second possible direction, the space of parameters can be studied using additional geometric techniques such as topological data analysis. 
In our experiments we observe that the dimension of the subspace changes during training and could possibly be correlated with performance. This opens up to future directions of research of further exploring this correlation. We hypothesize that if the dimensionality of the subspace is known a priori, developing a regularizer that enforces this could possibly improve performance for GBML methods. 

\section{Acknowledgements}
We thank Giovanni Luca Marchetti, Vladislav Polianskii and Marco Moletta for useful discussions. This work has been supported by the European Research Council (BIRD: 884807), Swedish Research Council and Knut and Alice Wallenberg Foundation and H2020 CANOPIES.


\bibliography{example_paper}
\bibliographystyle{icml2022}

\newpage
\appendix
\onecolumn
\section{Appendix}

\subsection{Classification}
\begin{figure*}[h]
    \centering
    \includegraphics[width=1.\textwidth]{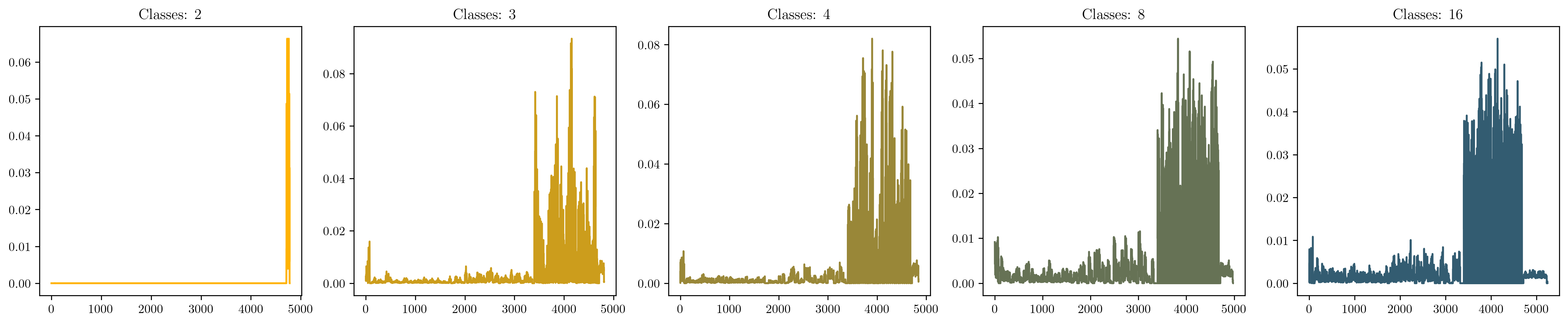}
    \caption{Mean absolute value of difference between parameters in the toy classification problem. The change in parameters is concentrated to the later layers. We note that for $N = 2$, only a small set of parameters are altered during adaptation.}
    \label{fig:mean_absolute_parameter_change}
\end{figure*}

In figure \ref{fig:mean_absolute_parameter_change}, we empirically investigate the distribution of how the parameters change with different tasks. We take all the tasks in the test set and compute the mean absolute difference between all the parameters. This gives an indication of what parameters change between tasks. As can be seen in the figure, for classification, mostly the parameters in the later layers get updated which confirms the results of \cite{raghu2019rapid}. 

\subsection{Sine-Experiment}

\begin{figure}[h]
    \centering
    \includegraphics[width=0.7\textwidth]{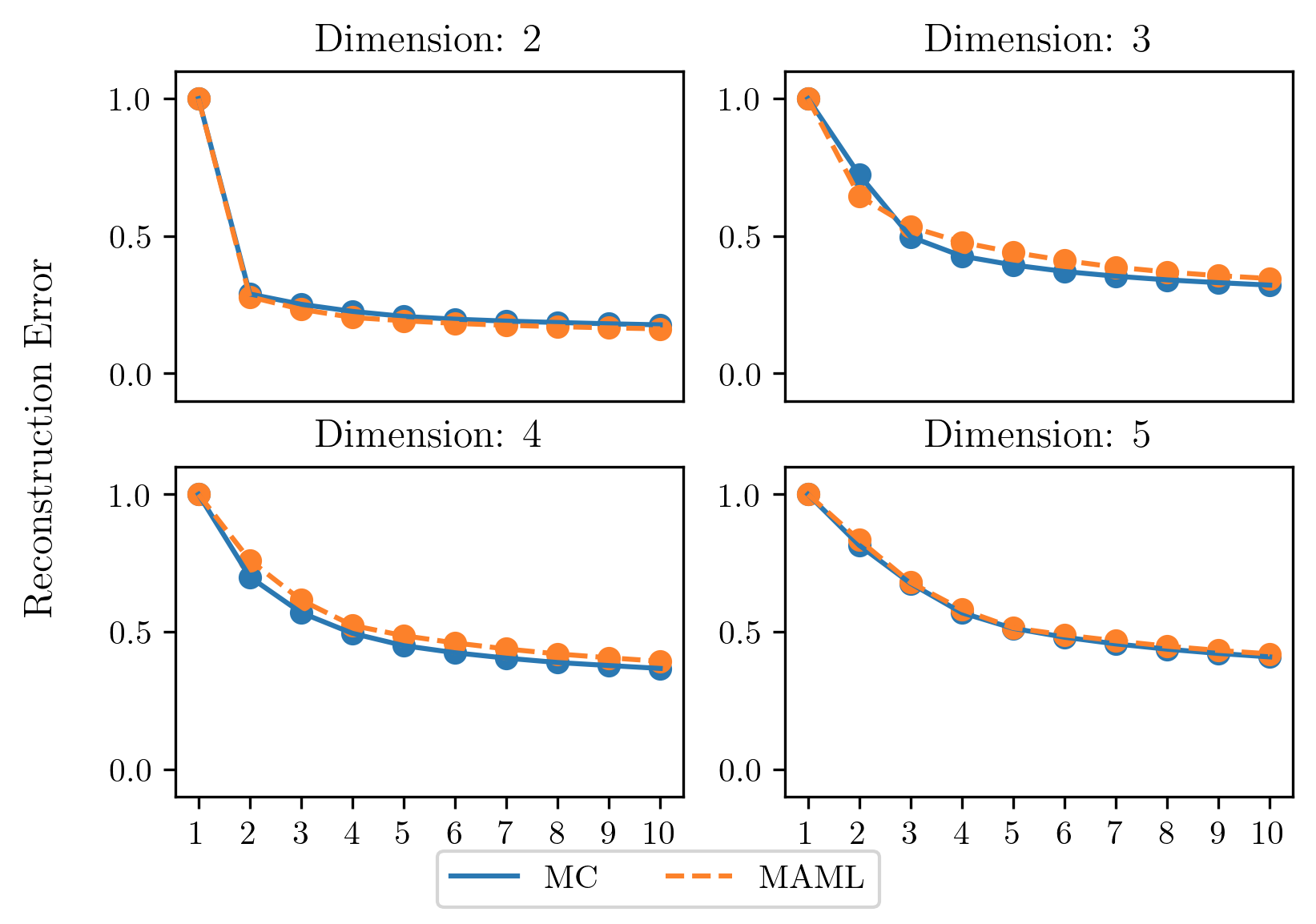}
    \caption{Isomap results for the sine experiment. }
    \label{fig:sine_isomap}
\end{figure}

\subsection{Omniglot}
\begin{figure}[h]
    \centering
    \includegraphics[width=0.7\textwidth]{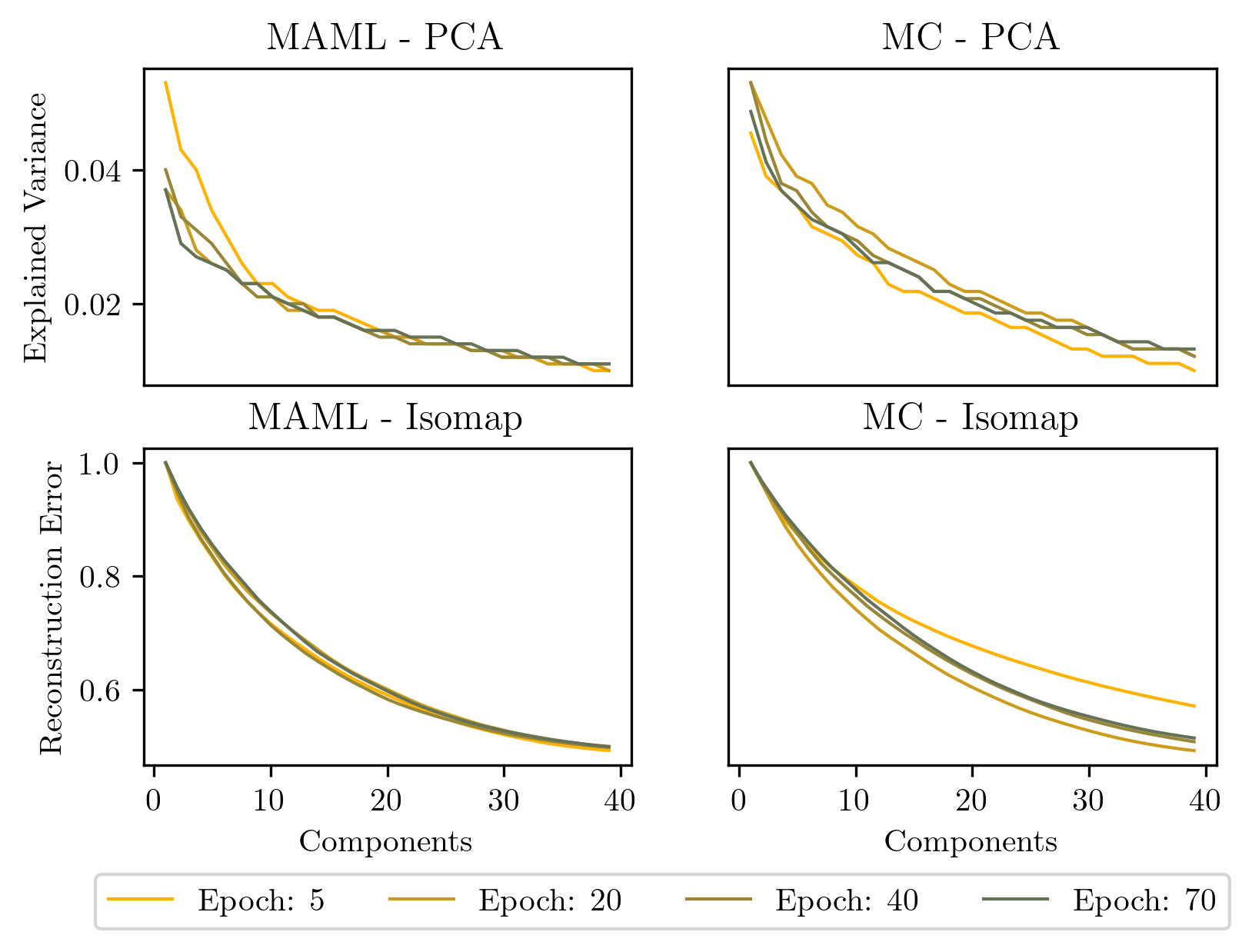}
    \caption{Omniglot results for PCA and Isomap.}
    \label{fig:omniglot_results}
\end{figure}

We also investigate few-shot classification on Omniglot. The results are depicted in figure \ref{fig:omniglot_results}. 

\subsection{miniImagenet}

\begin{figure}[h]
    \centering
    \includegraphics[width=0.7\textwidth]{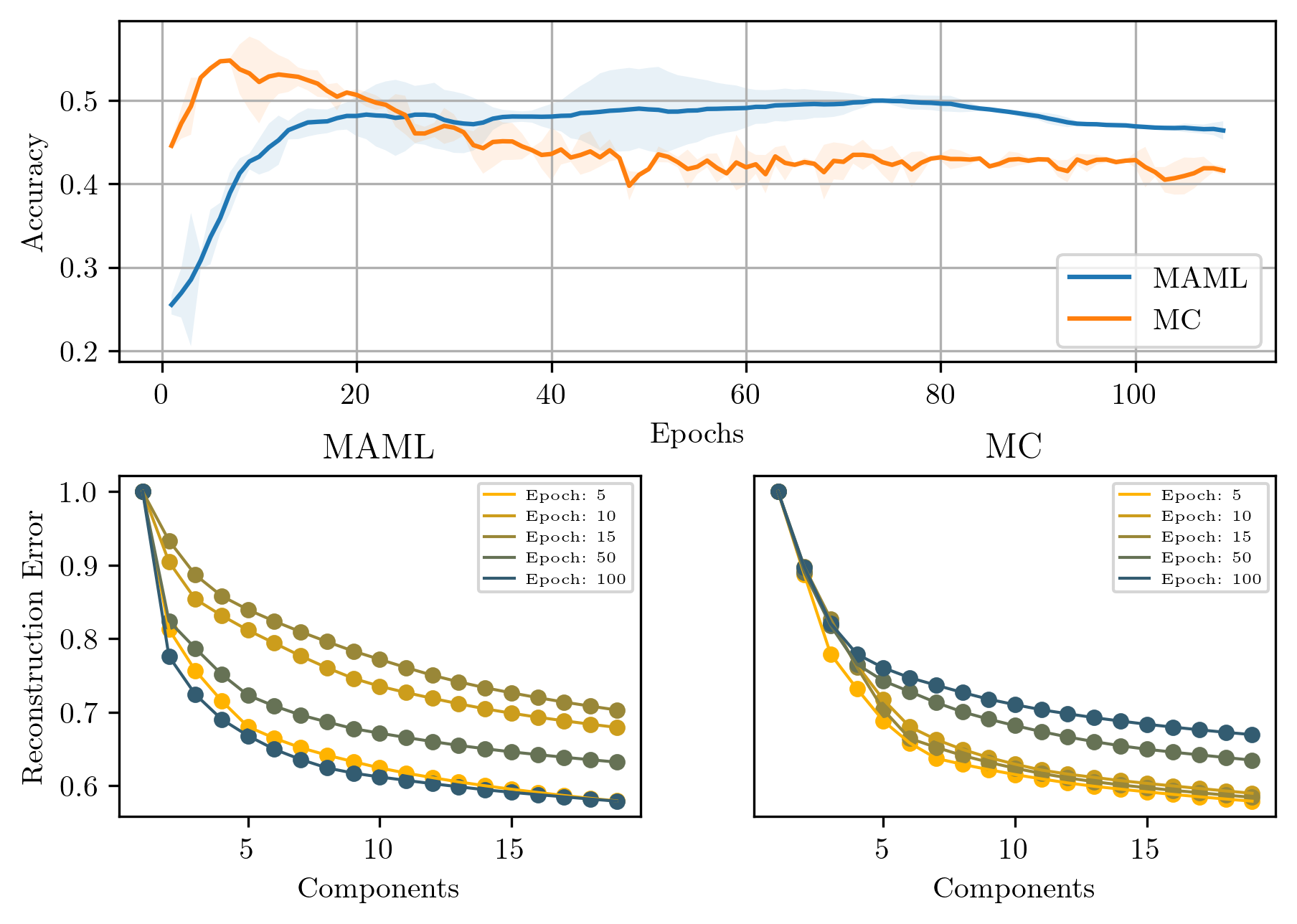}
    \caption{Performance of MAML and Meta-Curvature on a test-set of miniImagenet. Meta-Curvature begins to overfit after epoch $15$ while MAML shows increasing performance. }
    \label{fig:acc_mini_imagenet}
\end{figure}

In Figure \ref{fig:acc_mini_imagenet}, we provide a comparison between the estimated dimension and the test-set performance of the models at different stages during the training. The estimated dimensionality varies during training with MAML seemingly starting off with a low-dimensional parameter space which grows larger in the first epochs then settles down again as performance increases. For meta-curvature, we can observe a clear overfitting on the test set with performance peaking at around epoch $10$. This is reflected in the dimensionality estimation as Isomap tends to estimate a lower dimension for later epochs. 

\newpage

\subsection{Reconstruction Experiment - Toy Classification}

\begin{figure*}[h]
    \centering
    \includegraphics[width=0.7\textwidth]{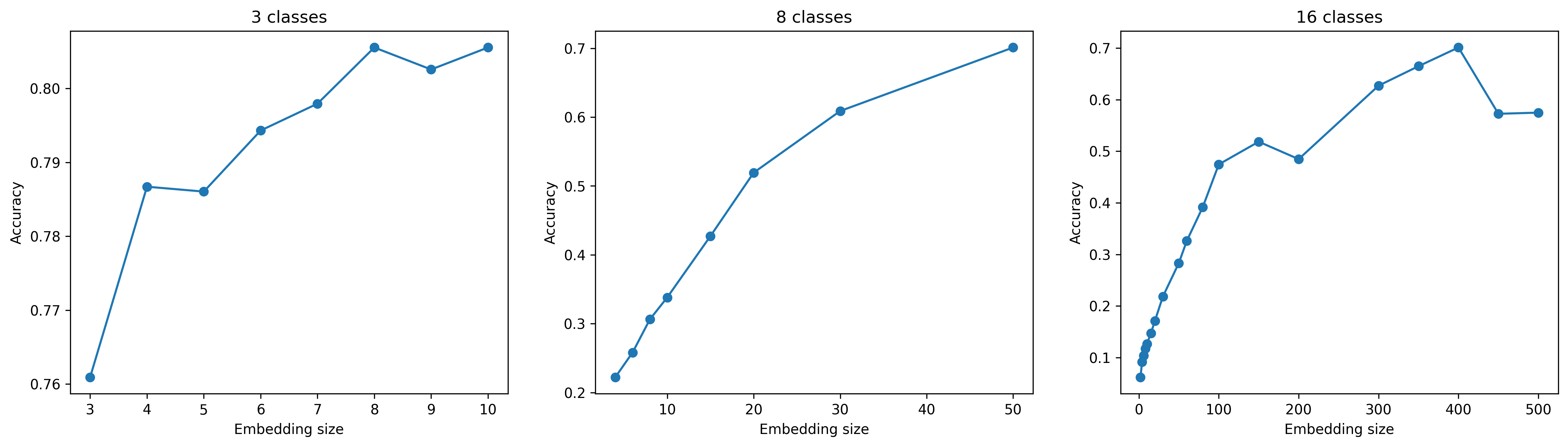}
    \caption{Accuracy of a trained classifier compared to embedding size for the toy-classification experiment.}
    \label{fig:toy_classification_reconstruction}
\end{figure*}


\end{document}